\pdfoutput=1

\documentclass[11pt]{article}

\usepackage{acl}

\usepackage{times}
\usepackage{bm}
\usepackage{latexsym}
\usepackage{pifont}

\usepackage[T1]{fontenc}

\usepackage[utf8]{inputenc}

\usepackage{microtype}

\usepackage{inconsolata}

\usepackage{amsmath}
\usepackage{amsfonts}
\usepackage{bm}
\usepackage{hyperref}
\usepackage{booktabs}
\usepackage{multirow}
\usepackage{diagbox}
\usepackage{tabularx}
\usepackage{arydshln}
\usepackage{enumitem}
\usepackage{CJKutf8}

\usepackage{graphicx}
\title{Bridging the Gap between Different Vocabularies for LLM Ensemble}

\author{
    Yangyifan Xu\textsuperscript{1,2} \thanks{\ \ Equal Contribution}, \ 
    Jinliang Lu\textsuperscript{1,2} \footnotemark[1], \ 
    Jiajun Zhang\textsuperscript{1,2,3,4}\thanks{\ \ Corresponding Author}   \\
    \textsuperscript{1}School of Artificial Intelligence, University of Chinese Academy of Sciences\\
    \textsuperscript{2}Institute of Automation, Chinese Academy of Sciences\\
    \textsuperscript{3}Wuhan AI Research, 
    \textsuperscript{4}Shanghai Artificial Intelligence Laboratory, Shanghai, China\\
    \texttt{\{xuyangyifan2021, lujinliang2019\}@ia.ac.cn}, \texttt{jjzhang@nlpr.ia.ac.cn} \\
}

\begin{document}
\maketitle
\begin{abstract}
Ensembling different large language models (LLMs) to unleash their complementary potential and harness their individual strengths is highly valuable. Nevertheless, vocabulary discrepancies among various LLMs have constrained previous studies to either selecting or blending completely generated outputs. This limitation hinders the dynamic correction and enhancement of outputs during the generation process, resulting in a limited capacity for effective ensemble. To address this issue, we propose a novel method to \textbf{E}nsemble LLMs via \textbf{V}ocabulary \textbf{A}lignment (EVA). EVA bridges the lexical gap among various LLMs, enabling meticulous ensemble at each generation step. Specifically, we first learn mappings between the vocabularies of different LLMs with the assistance of overlapping tokens. Subsequently, these mappings are employed to project output distributions of LLMs into a unified space, facilitating a fine-grained ensemble. Finally, we design a filtering strategy to exclude models that generate unfaithful tokens. Experimental results on commonsense reasoning, arithmetic reasoning, machine translation, and data-to-text generation tasks demonstrate the superiority of our approach compared with individual LLMs and previous ensemble methods conducted on complete outputs. Further analyses confirm that our approach can leverage knowledge from different language models and yield consistent improvement.\footnote{Our code is available in \url{https://github.com/xydaytoy/EVA}}
\end{abstract}
\section{Introduction}
Large language models (LLMs) have demonstrated impressive performance across various natural language processing tasks~\citep{anil2023palm,touvron2023llama,vicuna2023}. These models, spanning diverse datasets, architectures, and training methodologies, exhibit different strengths and weaknesses~\citep{llm-blender-2023}. Therefore, ensembling these LLMs to unleash their complementary potential and harness their individual strengths is highly valuable~\citep{llm-blender-2023,lu2023routing,shnitzer2023large}.

\begin{figure}[t]
    \centering
    \includegraphics[width=\hsize]{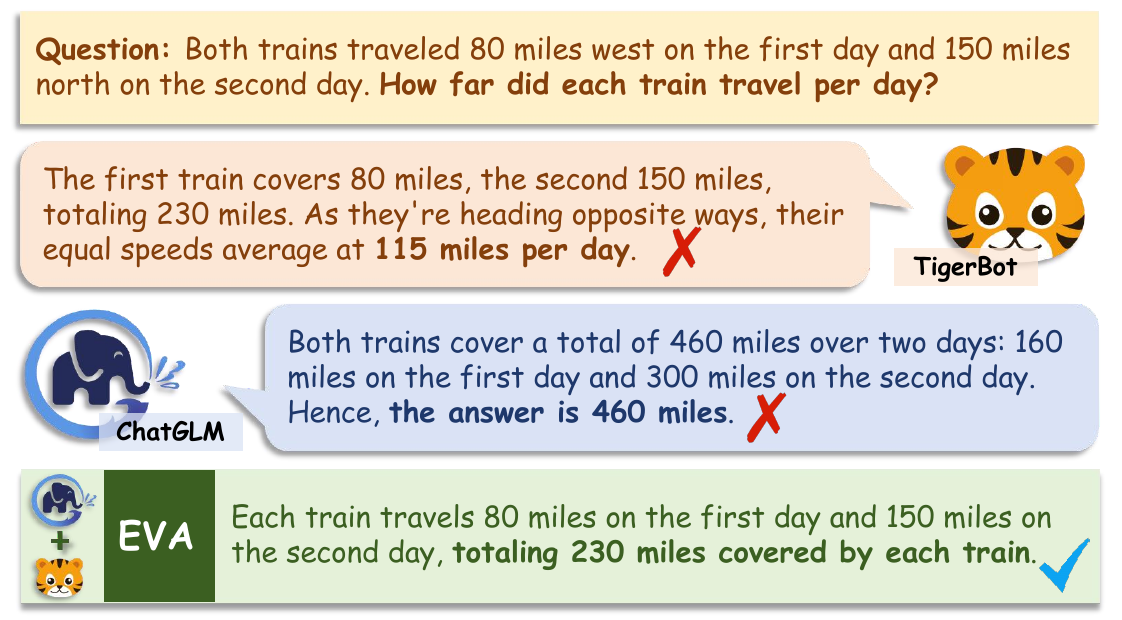}
    \caption{\textbf{Motivation of EVA.} For the problem of \textit{train travel distance}, both TigerBot and ChatGLM provide wrong answers. Ensembling over completely generated outputs cannot derive the correct answer. EVA achieves correct answers by performing fine-grained ensemble at each generation step, allowing each token to benefit from the ensemble.}
    \label{fig.motivation}
\end{figure}

\begin{figure*}[ht!]
    \centering
    \includegraphics[width=\textwidth]{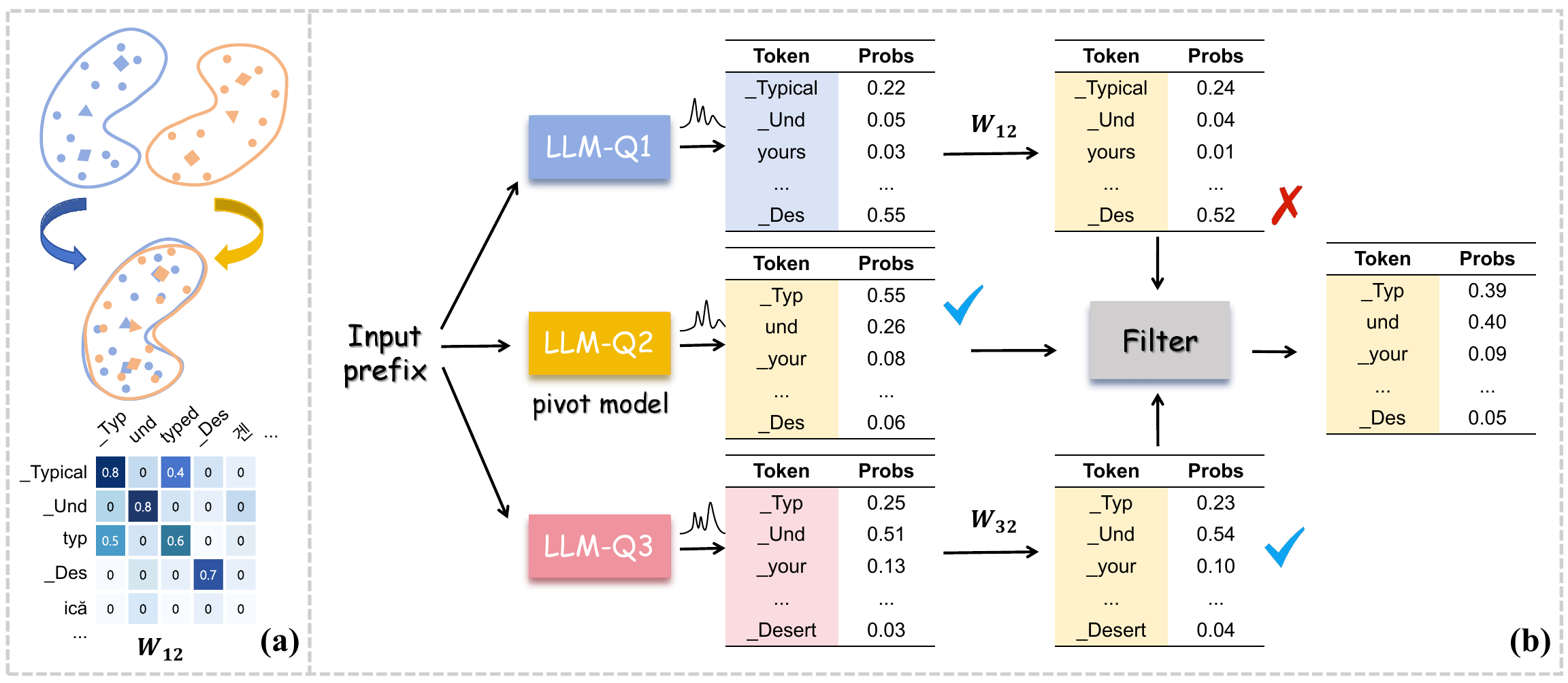}
    \caption{\textbf{The EVA framework.} EVA consists of two steps. (a) Firstly, we establishes alignment between the vocabularies of different models. (b) Next, we project the output distributions of different LLMs into a unified space using the established vocabulary alignment and exclude unfaithful tokens to perform fine-grained ensemble.}
    \label{fig.framework}
\end{figure*}

Previous studies typically concentrate on the ensemble of completely generated outputs, which involve either ranking multiple outputs to select the best one~\cite{lu2023routing,shnitzer2023large} or incorporating additional fusion models to blend these outputs~\cite{llm-blender-2023}. Therefore, these methods usually lead to ensemble outcomes confined to the space of several completely generated outputs. As shown in Figure~\ref{fig.motivation}, for the problem of \textit{train travel distance}, both TigerBot and ChatGLM provide incorrect reasoning processes, resulting in wrong answers. Ensembling over completely generated outputs cannot produce correct answer if all the candidate complete outputs are wrong.

One potential solution to this problem involves incorporating ensembling into the generation process of LLMs. As indicated by \citet{zhang2023language}, early errors in LLMs tend to snowball, leading to subsequent errors that might not have otherwise occurred. Ensembling during generation helps prevent the generation of inaccurate tokens at each step, thereby reducing misleading cues for subsequent token generation. However, such an ensemble approach is unfeasible for LLMs due to vocabulary discrepancies. As illustrated in Figure~\ref{fig.framework}, the three LLMs use distinct vocabularies, leading to different output distributions over tokens. This divergence hinders the straightforward token-level ensemble at each generation step.

To tackle this issue, we propose a simple yet effective method named \textbf{E}nsemble via \textbf{V}ocabulary \textbf{A}lignment (EVA), facilitating the fine-grained ensemble of LLMs at each generation step. EVA stems from a straightforward observation: although various LLMs have distinct vocabularies, they commonly share a significant number of overlapping tokens. By leveraging these tokens as bridges, EVA can achieve vocabulary alignment. Specifically, for vocabularies $\mathcal{V}^{Q_1}$, $\mathcal{V}^{Q_2}$ used in LLM-$Q_{1}$ and LLM-$Q_{2}$, we first extract embeddings of the overlapping tokens and learn a mapping matrix to project these embeddings into a shared space. Subsequently, by computing similarity scores between tokens in these vocabularies, we derive the semantic projection $\bm{W} \in \mathbb{R}^{|\mathcal{V}^{Q_1}| \times |\mathcal{V}^{Q_2}|}$. This enables the projection of output distributions from LLM-$Q_{1}$ to LLM-$Q_{2}$ and generates reasonable tokens based on the fused distribution of these LLMs at each inference step. Finally, we further enhance our approach by devising a filtering strategy capable of excluding models that generate unfaithful tokens.

Our method successfully overcomes the vocabulary discrepancy between different LLMs and facilitates fine-grained ensemble during generation. Significantly, our method necessitates solely an additional projection matrix $\bm{W}$, eliminating the necessity of extra fusion models or supervised training corpora.
We evaluate our method on various NLP tasks, including Commonsense Reasoning, Arithmetic Reasoning, Machine Translation, and Data-to-Text Generation. Experimental results demonstrate the superiority of our approach compared with individual LLMs and previous ensemble methods conducted on complete outputs. Further analyses confirm that our approach can leverage knowledge from different language models and yield consistent improvement.

Briefly, our contributions can be summarized as follows:
\begin{itemize}
    \item We propose a novel LLM ensemble method to achieve fine-grained ensemble at each generation step. Our method aims to bridge the lexical gap between LLMs, thereby unleashing their complementary potentials.
    \item We devise an effective filtering strategy to exclude models generating unfaithful tokens, preventing underperforming models from misleading the overall judgment.
    \item Empirical results demonstrate the effectiveness and superiority of our method, which significantly improves overall performance on various natural language processing tasks.
\end{itemize}

\section{Vocabulary Overlap Phenomenon}\label{sec:method1}

\subsection{Impact of Vocabulary Distinction}
Current LLMs accomplish various tasks through language generation, where LLMs receive the input prompt and generate succeeding tokens. Suppose the input tokens are $x_{1}, \cdots, x_{i-1}$, LLMs decode the next token $x_{i}$ based on the conditional distribution $p(\cdot | x_{\leq i}) \in \mathbb{R}^{|V|}$ over the corresponding vocabulary.

However, different LLMs usually independently learn sentencepiece \cite{kudo-richardson-2018-sentencepiece} models from different training corpora, leading to different vocabularies. For instance, the vocabulary size of LLaMA is 32,000, whereas ChatGLM has a vocabulary length of 125,696. Such a discrepancy makes the output distributions of different models noncomparable, thereby impeding direct ensembling, as commonly practiced in conventional classification tasks.

\subsection{Overlap between Vocabularies}
\begin{figure}[!ht]
\centering
\includegraphics[width=\hsize*5/6]{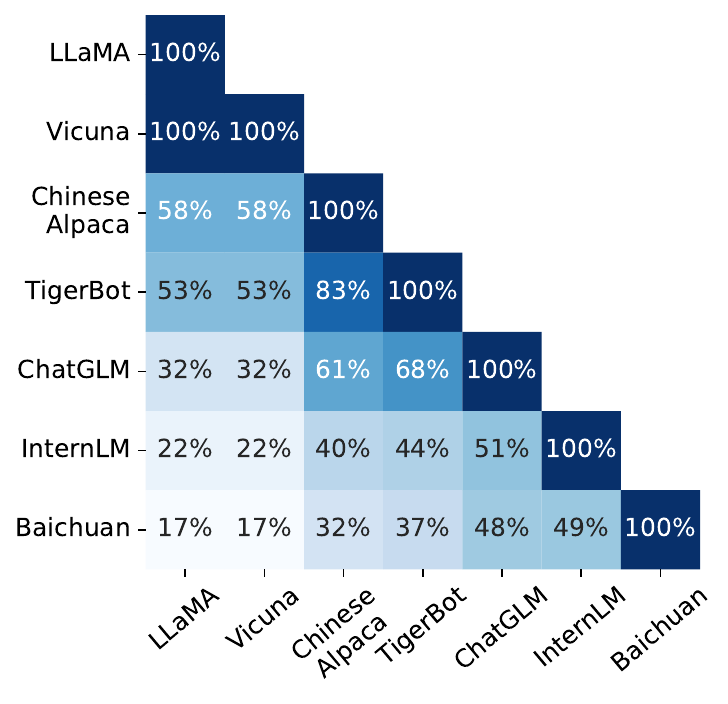}
\caption{The rate of overlapping tokens between different LLMs vocabularies. The models are arranged in ascending order based on vocabulary size. Each cell represents the proportion of shared tokens between the horizontal and vertical models, relative to the vocabulary size of the vertical model.}
\label{fig:identity_vocab}
\end{figure}

Although different LLMs have distinct vocabularies, given that these diverse vocabularies are learned from comparable corpora collected from the web, a substantial number of overlapping tokens naturally emerge. To illustrate this phenomenon, we record the rate of overlapping tokens between vocabularies of LLMs. As shown in Figure~\ref{fig:identity_vocab}, the number of overlapping tokens is adequate. For example, TigerBot and LLaMA have 53\% overlapping tokens.
Intuitively, these overlapping tokens play a crucial role as a bridge to project diverse output distributions into a shared space and establish the corresponding relations, facilitating the ensemble of LLMs. 

\section{Our Method}\label{sec:method}
EVA comprises two key components: \textit{cross-model vocabulary alignment} (Section~\ref{sec:cmva}) and \textit{LLMs ensemble} (Section~\ref{sec:method4}). The framework is shown in Figure~\ref{fig.framework}, \textit{(a) cross-model vocabulary alignment} establishes the relations between tokens of distinct vocabularies. \textit{(b) LLMs ensemble} projects the output distributions into the same space via the established vocabulary relations and achieves fine-grained ensembling at each generation step.

Considering a set of $N$ large language models denoted as $\mathcal{M}=\{Q_1,Q_2,\cdots,Q_{N-1},P\}$, where $P$ represents the chosen pivot model. We empirically select the model with the largest vocabulary as the pivot model $P$.\footnote{Please refer to the appendix~\ref{sec:app3} for details.}

\subsection{Cross-Model Vocabulary Alignment}\label{sec:cmva}

\subsubsection{Vocabulary Projection}\label{sec:method2}
As shown in the upper part of Figure~\ref{fig.framework}(a), We first utilize the overlapping tokens as supervised labels to map token embeddings from different models to a common vector space. 
Taking $N=2$ as an example, let $\mathcal{V}^{P}$ and $\mathcal{V}^{Q}$ represent the vocabularies of the pivot model and the non-pivot model, and $\bm{E}^{P}$ and $\bm{E}^{Q}$ be the word embedding matrices of the respective models. The training objective is to find transformation matrices $\bm{U}_{QP}$ such that:
\begin{equation}
\bm{U}_{QP}=\underset{\bm{U}_{QP}}{\operatorname{argmin}} \sum_i \sum_j \mathcal{D}_{i j}\left\|\bm{E}^{Q}_{i *} \ \bm{U}_{QP}-\bm{E}^{P}_{j *}\right\|^2
\end{equation}
where $\mathcal{D}$ is the overlapping dictionary of $\mathcal{V}^{Q}$ and $\mathcal{V}^{P}$, and $\mathcal{D}_{i j} = 1$ indicates that the $i$-th word in $\mathcal{V}^{Q}$ and the $j$-th word in $\mathcal{V}^{P}$ are identical.
We utilize the supervised setting of the open-source toolkit VecMap\footnote{\url{https://github.com/artetxem/vecmap}} to achieve the training process. This involves applying normalization, whitening, orthogonal mapping, re-weighting, and de-whitening operations to the word embeddings~\citep{artetxe2018generalizing}. The optimal $\bm{U}_{QP}$ minimizes the Euclidean distance between identical words from different model vocabularies in the mapped common space.

Subsequently, we establish vocabulary mappings between models based on the similarity relationships between tokens:
\begin{equation}
\bm{W}^{QP} = \operatorname{SIM}\left(\bm{E}^{Q} \bm{U}_{QP}, \bm{E}^{P}\right)
\end{equation}

Specifically, we adopt the cross-domain similarity local scaling (CSLS)~\citep{lample2018word} score as the token similarity from $\mathcal{V}^{Q}$ to $\mathcal{V}^{P}$ and derive the similarity matrix $\bm{W}^{QP} \in \mathbb{R}^{|\mathcal{V}^{Q}| \times |\mathcal{V}^{P}|}$.

\subsubsection{Noise Reduction}\label{sec:method3}
Since the similarity matrix obtained above is excessively large and contains substantial noise, we calculate the alignment across various similarity intervals (as shown in Table~\ref{tab.alignment}, with detailed analysis in Appendix~\ref{sec:app1}) and devise three steps to reduce noise and retain the pertinent and concise alignment information.

\paragraph{Step-1: Top-$t$ Truncation.}
The complete similarity matrix is redundant, as each token should only align with a small subset of other tokens. Thus, for each token in $\mathcal{V}_{Q}$, we retain top-$t$ tokens in $\mathcal{V}_{P}$ that exhibit the highest similarity to it.

\begin{equation}
\bm{W}_{ij}^{QP}= \begin{cases}\bm{W}_{ij}^{QP}, & \bm{W}_{ij}^{QP} \in \operatorname{top-}t\left(\bm{W}_{i*}^{QP}\right) \\ 0 & \text { otherwise }\end{cases}
\label{eq:rule1}
\end{equation}

\paragraph{Step-2: Threshold Truncation.}
When the similarity between two tokens is too low, aligning them becomes meaningless. Therefore, we set a threshold to discard the portion of similarity scores that are below the threshold.
\begin{equation}
\bm{W}_{ij}^{QP}= \begin{cases}\bm{W}_{ij}^{QP}, & \bm{W}_{ij}^{QP} \geq threshold \\ 0 & \text { otherwise }\end{cases}
\label{eq:rule2}
\end{equation}

\paragraph{Step-3: Variance Truncation.}
Through the observation of Table~\ref{tab.alignment}, we found that tokens without actual meaning exhibit similar and high similarity scores with multiple tokens, which cannot represent the semantic similarity. We use variance to determine and eliminate this noise, taking into account the number of non-zero similarity scores as well to avoid low variance resulting from insufficient quantity.

\begin{small}
\begin{equation}
\bm{W}_{ij}^{QP}\!=\!\begin{cases}0, & 
\operatorname{Var}\left(\bm{W}_{i*}^{QP}\right)\!\leq\!\sigma, \operatorname{count}\left(\bm{W}_{i*}^{QP}\!\neq\!0\right)\!\geq\!c
\\ \bm{W}_{ij}^{QP} & \text { otherwise }\end{cases}
\label{eq:rule3}
\end{equation}
\end{small}

Following these three processes, we obtain a sparse and efficient mapping matrix $\bm{W}^{QP}$, which is only about 1MB. This matrix maps the output distribution of the non-pivot model to the pivot model.

\begin{table}[!t]
    \centering
    \includegraphics[width=\hsize]{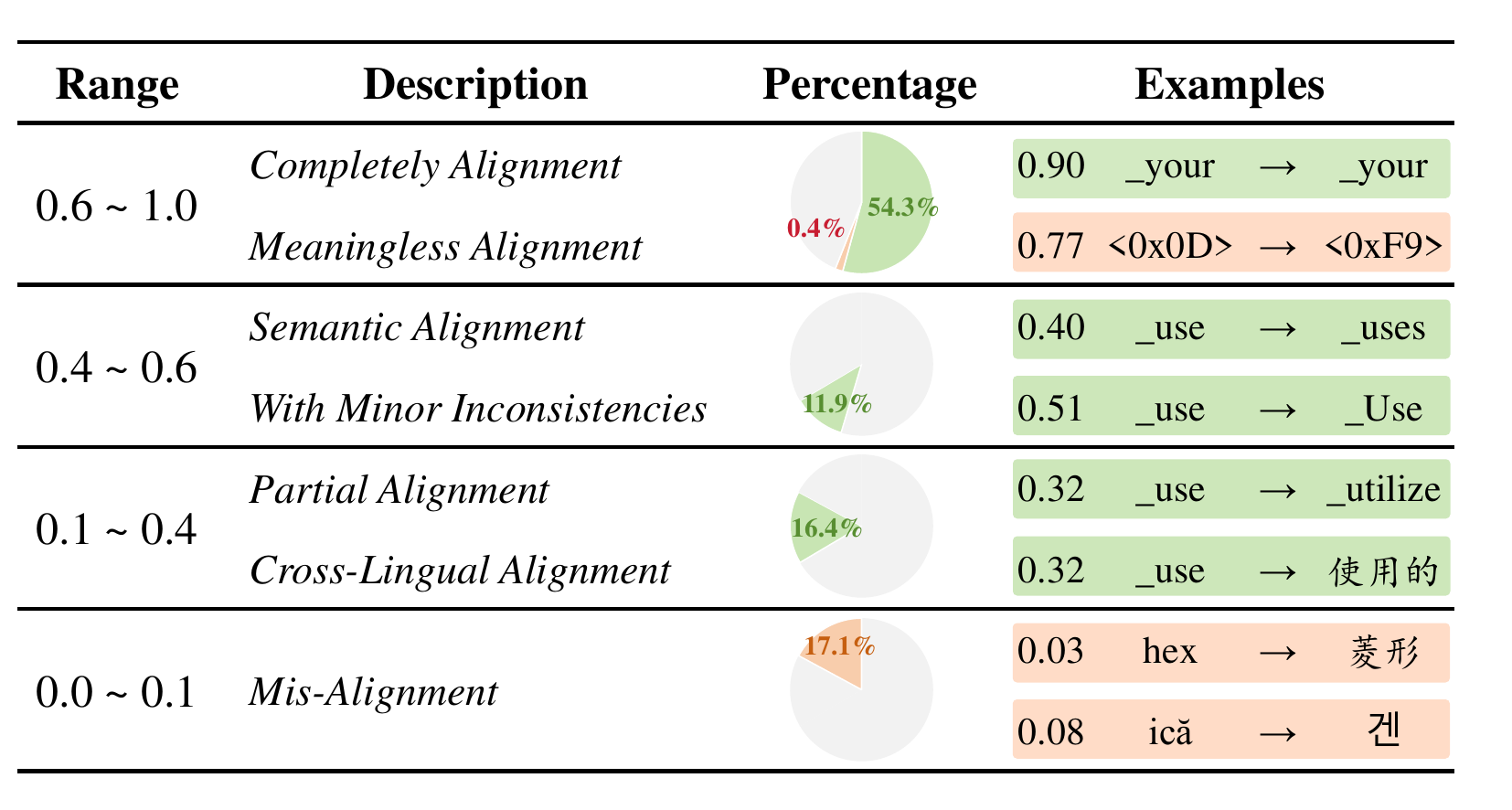}
    \caption{Statistics of token alignment from LLaMA to Baichuan. Similarity scores are divided into four subsets based on alignment performances. We intend to retain the pairs highlighted in green and discard those highlighted in red.}
    \label{tab.alignment}
\end{table}

\subsection{LLMs Ensemble}\label{sec:method4}
As shown in Figure~\ref{fig.framework}(b), given the mapping matrix (e.g., $\bm{W}^{12}$ and $\bm{W}^{32}$) from non-pivot models ($Q_{1}$ and $Q_{3}$) to the pivot model ($Q_{2}$), we align the output distribution of non-pivot models at the current time step with the pivot model.
\begin{equation}
p_{\ell} \left(\cdot \mid x_{<i}\right) = q_{\ell} \left(\cdot \mid x_{<i}\right)  \bm{W}^{\ell \rho} \quad \forall \ell \neq \rho.
\end{equation}
where $\rho$ is the identifier for the pivot model, $q_{\ell}\left(\cdot \mid x_{<i}\right)$ and $p_{\ell}\left(\cdot \mid x_{<i}\right)$ separately denote the original output distribution of the $\ell$-th model in $\mathcal{M}$ and its corresponding mapping in the unified space.

A straightforward ensemble approach involves deriving the succeeding token by averaging the mapped output distributions of all models:
\begin{equation}
    p \left(\cdot \mid x_{<i}\right) = \frac{1}{N} \sum_{\ell=1}^N p_{\ell} \left(\cdot \mid x_{<i}\right)
\end{equation}

\begin{table*}[!ht]
\centering
\fontsize{10.0}{12.0} \selectfont
\resizebox{\textwidth}{!}{
\begin{tabular}{@{}lccccc@{}}
\toprule
\multirow{2}{*}{\textbf{}} & \multicolumn{4}{c}{\textbf{Machine Translation}}                                                      & \textbf{Data-to-Text}          \\ 
                           & \multicolumn{2}{c}{\textbf{Flores-Zh-En}}     & \multicolumn{2}{c}{\textbf{Flores-En-Zh}}    & \textbf{E2E}                   \\ \cmidrule(l){2-3} \cmidrule(l){4-5} \cmidrule(l){6-6} 
\textbf{System}            & BLEU                  & ChrF                  & BLEU                  & ChrF                 & ROUGE-L               \\ \midrule
LLaMA2-7B-Chat             & 24.49                 & 52.37                 & 13.99                 & 22.78                & 33.58                 \\
ChatGLM2-6B                & 24.17                 & 51.71                 & \underline{23.77}           & 31.14                & \underline{40.57}           \\
Baichuan2-7B-Chat          & \underline{29.18}           & 56.63                 & \underline{30.56}           & 35.95                & 30.61                 \\
InternLM-7B-Chat           & 22.59                 & 51.81                 & 23.58                 & 31.18                & \underline{41.11}           \\
TigerBot-7B-Chat-V3        & \underline{26.81}           & 54.34                 & \underline{30.59}           & 35.58                & 20.37                 \\
Vicuna-7B-V1.5             & \underline{26.37}           & 53.83                 & 20.61                 & 28.98                & \underline{37.08}           \\
ChineseAlpaca2-7B          & \underline{28.54}           & 54.42                 & \underline{27.66}           & 33.87                & \underline{38.24}           \\ \midrule
MBR~\cite{farinhas2023empirical}                        & 30.72(\textit{+1.54})          & 56.97(\textit{+0.34})          & 31.29(\textit{+0.70})           & 36.84(\textit{+0.89})         & 41.47(\textit{+0.36})          \\
PairRanker~\citep{llm-blender-2023}                 & 29.73(\textit{+0.55})          & 56.58(\textit{-\ 0.05})          & 29.45(\textit{-\ 1.41})           & 35.25(\textit{-\ 0.70})         & 38.90(\textit{-\ 2.21})          \\
LLM-Blender~\citep{llm-blender-2023}                 & 27.18(\textit{+1.54})          & 53.89(\textit{+0.34})          & -          & -         & \textbf{43.62(\textit{+2.51})}          \\ \midrule
EVA (\textit{ours})          & \textbf{31.16(\textit{+1.98})}          & \textbf{57.77(\textit{+1.14})}          & \textbf{32.68(\textit{+2.09})}          & \textbf{38.16(\textit{+2.21})}        & 42.62(\textit{+1.51}) \\ \bottomrule
\end{tabular}}
\caption{\label{table:mt_result} Main results of machine translation and data-to-text tasks. Best results are highlighted in bold and the model employed within the ensemble is underlined for distinction. LLM-Blender is not trained on Chinese corpora, thus unable to produce meaningful translations from English to Chinese.}
\end{table*}
However, this approach is susceptible to outliers, which can mislead overall judgments. 
Hence, we devise a filtering strategy to enforce a requisite consistency among tokens generated by diverse models.
Specifically, if the top-1 token predicted by a model falls outside the top-$n$ tokens predicted by any other model, it is excluded from the ensemble. 
\begin{equation}
    p\!\left(\cdot\!\mid\!x_{<i}\right)\!=\!\frac{1}{\sum_{\ell=1}^N\!I\left(\ell\right)} \sum_{\ell=1}^N I\left(\ell\right)\!\cdot\!p_{\ell}\!\left(\cdot\!\mid\!x_{<i}\right)
\end{equation}
\begin{equation}
    I\left(\ell\right)\!=\!\begin{cases}
        1 & \text{if } \operatorname{top-1}\left(p_{\ell}\right) \in \bigcup\limits_{\substack{o \neq \ell}} \operatorname{top-}n\left(p_{o}\right) \\
        0 & \text{otherwise}
    \end{cases} 
\end{equation}

As shown in Figure~\ref{fig.framework}(b), When we directly average the probability distributions of the three models, the ensemble result is \textit{\_Typ}. Upon incorporating the filtering strategy with $n=3$, the top-1 token for model $Q_{1}$ is \textit{\_Des}, which is not within the top-3 tokens of $Q_{2}$ or $Q_{3}$, hence excluded from ensemble. 
On the contrary, the top-1 token of $Q_{2}$ is \textit{\_Typ}, falling within the top-3 tokens of $Q_{1}$ and $Q_{3}$. The top-1 token of $Q_{3}$ is \textit{und}, within the top-3 tokens of $Q_{2}$. Consequently, we ensemble only Q2 and Q3, resulting in the correct output \textit{und}.

\section{Experimental Settings}\label{sec:experiments}

\subsection{Datasets}
We evaluate our proposed ensemble method from the perspective of natural language generation (NLG) and reasoning. For NLG, we choose machine translation (Flores-101 Chinese$\leftrightarrow$English) \cite{flores101} and data-to-text generation task (E2E) \cite{novikova2017e2e}. For commonsense reasoning, we employ Natrual Question (NQ) \cite{kwiatkowski2019natural} and TriviaQA \cite{joshi2017triviaqa} for evaluation. For arithmetic reasoning, we adopt GSM8K \cite{cobbe2021gsm8k}, AddSub \cite{hosseini2014learning} and ASDiv \cite{miao2020diverse} for evaluation.\footnote{Please refer to the appendix~\ref{sec:app2} for details of the tasks.}

\subsection{Candidate LLMs}
We select seven open-source chat LLMs of approximately 7B size as the candidate LLMs for the ensemble as follows: LLaMA2-7B-Chat~\citep{touvron2023llama}, ChatGLM2-6B~\citep{zeng2022glm}, Baichuan2-7B-Chat~\citep{baichuan2023baichuan2}, InternLM-7B-Chat~\citep{2023internlm}, TigerBot-7B-Chat-V3~\citep{chen2023tigerbot}, Vicuna-7B-V1.5~\citep{vicuna2023}, ChineseAlpaca2-7B~\citep{Chinese-LLaMA-Alpaca}.\footnote{Integration of models of different sizes is discussed in appendix~\ref{sec:app4}.}

These models originate from distinct institutions and have different vocabularies. Each model is aligned by supervised instruction tuning and leverages large-scale, high-quality data to establish a powerful knowledge base, thus performing well on public benchmarks.

\begin{table*}[!ht]
\centering
\fontsize{10.0}{11.5} \selectfont
\resizebox{\textwidth}{!}{
\begin{tabular}{@{}lccccc@{}}
\toprule
 & \multicolumn{2}{c}{\textbf{Commonsense Reasoning}}                       & \multicolumn{3}{c}{\textbf{Arithmetic Reasoning}} \ \\ \cmidrule(l){2-3} \cmidrule(l){4-6}
\ \textbf{System}         & \textbf{NQ} & \textbf{TriviaQA}           & \textbf{GSM8K}  &  \textbf{AddSub}  &  \textbf{ASDiv}    \\    \midrule
\ LLaMA2-7B-Chat                   & \underline{28.59}            & \underline{62.77}        & 24.64    & \underline{55.05} & \underline{55.02}         \   \\
\ ChatGLM2-6B                      & 14.93                     & 31.77     & \underline{30.78}   & \underline{49.54} & \underline{60.52}       \      \\
\ Baichuan2-7B-Chat                & \underline{24.07}            & \underline{55.62}      & \underline{29.95}   & \underline{55.05}  &  \underline{58.74}      \     \\
\ InternLM-7B-Chat                 & 17.20                      & 44.05         & \underline{32.30}  & \underline{62.39} & \underline{58.58}          \     \\
\ TigerBot-7B-Chat-V3              & 11.33                     & 23.87      & \underline{27.29}   & 24.77 & 41.75         \    \\
\ Vicuna-7B-V1.5                   & \underline{26.84}            & \underline{61.21}       & 18.88   & 44.04 & 44.17                  \    \\
\ ChineseAlpaca2-7B                & \underline{22.58}            & \underline{50.86}                   & 13.12      & 23.85 & 28.64               \    \\ \midrule
\ MBR~\citep{farinhas2023empirical}                      & 28.61(\textit{+0.02})              & 63.75(\textit{+0.98})         & 36.47(\textit{+4.17})    & 58.72(\textit{-3.67}) &  61.00(\textit{+0.48})          \   \\
\ PairRanker~\citep{llm-blender-2023}                      & 29.81(\textit{+1.22})              & 63.24(\textit{+0.47})         & 39.58(\textit{+7.28})    & 58.72(\textit{-3.67}) & 62.62(\textit{+2.10})            \   \\
\ LLM-Blender~\citep{llm-blender-2023}                      & \textbf{32.19(\textit{+3.60})}              & 62.77(\textit{+0.00})         & 34.80(\textit{+2.50})    & 58.72(\textit{-3.67}) & 59.71(\textit{-0.81})            \   \\ \midrule
\ EVA (\textit{ours})                           & 30.64(\textit{+2.05})     & \textbf{64.29(\textit{+1.52})} & \textbf{42.91(\textit{+10.61})}   & \textbf{64.22(\textit{+1.83})} & \textbf{65.05(\textit{+4.53})}  \   \\ \bottomrule
\end{tabular}
}
\caption{\label{table:qa_result} 
Main results of commonsense reasoning (measured by Exact Match) and arithmetic reasoning tasks (measured by Accuracy). Best results are highlighted with bold and the model employed within the ensemble is underlined for distinction.}
\end{table*}

\subsection{Baselines}
We compare EVA with existing selection-based methods and fusion-based methods.
\paragraph{MBR} \citet{farinhas2023empirical} use the average similarity between one output and the rest to select the best output. We utilize BERTScore to measure the similarity between two outputs to adapt across different tasks. 
\paragraph{PairRanker} \citet{llm-blender-2023} employ a specialized pairwise comparison method to distinguish subtle differences between candidate outputs.
\paragraph{LLM-Blender} \citet{llm-blender-2023} utilize a 3b-parameter model fine-tuned on an instruction dataset to merge the ranking outcomes from PairRanker and generate the final output.

\subsection{Implement Details}

\paragraph{Configurations.} For each task, we selected the top-performing four models out of seven for the ensemble. We employ greedy decoding in all experiments since it generally produces higher-quality outputs. 
To mitigate the impact of long-tail noise accumulation, we perform top-$k$ truncation on the original output distributions of each candidate model.

\paragraph{Hyperparameters.} Unless otherwise stated, the same hyper-parameters are used in all experiments. 
Concerning the three steps mentioned in Section~\ref{sec:method2}, we empirically set $t=10$, $threshold=0.1$, $sigma=0.0001$ and $c=5$ based on observations. 
For top-$k$ truncation on the output distributions, we always set $k=320$ for the main results in the paper, which is quite robust across various tasks.  
Due to variations in task characteristics, we empirically set $n=40$ for NLG tasks and $n=3$ for reasoning tasks in our experiments.

\paragraph{Prompting.} 
For machine translation tasks, we utilize a 4-shot in-context learning setting, whereas for other tasks, we conduct zero-shot inference. Additionally, we include a chain of thought prompt in arithmetic reasoning tasks. We adhere to the specific format required by each chat model and employ task-specific prompts.

\begin{table*}[!h]
\resizebox{\textwidth}{!}{
\begin{tabular}{@{}lcccccccc@{}}
\toprule
 & \multicolumn{3}{c}{\textbf{Arithmetic Reasoning}} & \multicolumn{2}{c}{\textbf{\ \ Commonsense Reasoning}} & \multicolumn{2}{c}{\ \ \textbf{Machine Translation}} & \textbf{Data-to-Text} \\ \cmidrule(l){2-4}  \cmidrule(l){5-6}  \cmidrule(l){7-8} \cmidrule(l){9-9} 
{\color[HTML]{262626} \textbf{System}} &    \textbf{GSM8K} & \textbf{AddSub} & \textbf{ASDiv}  &  \ \ \ \ \ \  \textbf{NQ}  & \ \ \textbf{TriviaQA}   & \ \ \ \textbf{Zh-En}        & \ \ \ \textbf{En-Zh}       & \textbf{E2E}          \\ \midrule
$\text{EVA}_{n=40}$                                             & 31.39    &  58.72 &        61.33     &  \ \ \ \  \ \  30.86        & \ \ 64.59      & \ \ \ 31.16                   & \ \ \ 32.68                  & 42.62                 \\
$\text{EVA}_{n=20}$                                           & 31.54    &  59.63 &  60.68       &  \ \ \ \ \ \  30.61        & \ \ 64.48           & \ \ \ 31.20                   & \ \ \  32.78                  & 42.64                 \\
$\text{EVA}_{n=10}$                                          & 35.03    &  59.63 &  63.27    &  \ \ \ \ \ \  30.83        & \ \ 64.41                & \ \ \ 31.13                   & \ \ \  32.78                  & 42.59                 \\
$\text{EVA}_{n=5}$                                              & 37.30    &  62.39 &  65.86     &  \ \ \ \ \ \  30.75        & \ \ 64.26            & \ \ \ 31.01                   & \ \ \ 32.67                  & 42.00                 \\
$\text{EVA}_{n=3}$                                           & 42.91    &  64.22 &  65.05       &  \ \ \ \ \ \  30.64        &  \ \ 64.29             & \ \ \ 31.13                   & \ \ \ 32.64                  & 41.98                 \\ \bottomrule
\end{tabular}}
\caption{\label{table:ablation_result} 
Effect of model filtering intensity.}
\end{table*}

\section{Experimental Results}
The main results on NLG tasks and reasoning tasks are shown in Table~\ref{table:mt_result} and Table~\ref{table:qa_result}, respectively.\footnote{We conduct a human analysis of token alignment in appendix~\ref{sec:app5}.}
\paragraph{EVA demonstrates superiority.}
Our proposed EVA consistently outperforms individual models and selection-based ensemble methods across all types of tasks, showcasing its cross-task versatility. Especially in the GSM8K task, EVA achieves a significant 10.61 improvement compared with the best-performing individual model, ChatGLM2-6B.
Remarkably, EVA also outperforms LLM-Blender, which leverages an additional 3b-parameter fusion model, on six out of eight tasks, demonstrating the effectiveness of our approach.
We attribute this success to the EVA which conducts fine-grained ensembles at each generation step, ensuring precision in token generation and thereby mitigating subsequent errors in the generation of following tokens.

\paragraph{LLMs have diverse strengths and weaknesses.} Additionally, observing the performance of individual models on each task, we find that no models participate in every task ensemble. However, each model contributes to at least three task ensembles. This highlights the distinct knowledge possessed by each LLM and emphasizes the significance of ensembling LLMs.

\begin{table*}[]
\resizebox{\textwidth}{!}{
\begin{tabular}{@{}ll@{}}
\toprule[1.5pt]
\multicolumn{2}{c}{\textit{Flores-Zh-En}}                                                                                                                         \\ \midrule
Input           & \begin{CJK}{UTF8}{gbsn}他补充道：“我们现在有 4 个月大没有糖尿病的老鼠，但它们曾经得过该病。”\end{CJK}                                                                                                           \\ \cmidrule(l){2-2} 
Output prefix   & He added, "We have 4-month-                                                                                                                     \\ \cmidrule(l){2-2} 
Continuations & old mice that have never had diabetus, but they have had it in the past."                                            \\ \cmidrule(l){2-2} 
Next token distribution     & 'old', 'olds', ' old', 'Old', 'older', ' Old', 'OLD', ' olds', '\begin{CJK}{UTF8}{gbsn}旧\end{CJK}', 'olding', ...                                                              \\ \midrule
\multicolumn{2}{c}{\textit{GSM8K}}                                                                                                                                \\ \midrule
Input           & Janet\textbackslash{}u2019s ducks lay 16 eggs...How much in dollars does she make every day at the farmers' market?                             \\ \cmidrule(l){2-2} 
Output prefix   & First, we need to determine how many eggs Janet has left after she eats three for breakfast and bakes                                           \\ \cmidrule(l){2-2} 
Continuations & \ four muffins...The answer: 10. \\ \cmidrule(l){2-2} 
Next token distribution     & ' four', ' muff', ' the', ' some', ' ', ' a', ' three', ' her', ' for', ' two', ...                                                         \\ \bottomrule[1.5pt]
\end{tabular}}
\caption{\label{table:ana2} 
Examples of the distribution of the next token for GSM8K and Flores-Zh-En tasks.}
\vspace{-5pt}
\end{table*}

\section{Analysis}

\subsection{Effect of Model Filtering Intensity}\label{sec:4.1}

Recall in Section~\ref{sec:method4}, we introduced the hyperparameter $n$ as a way to control how strict our model filtering is. 
In this section, we investigate the sensitivity of our method to $n$.
As shown in Table~\ref{table:ablation_result}, all tasks, except for arithmetic reasoning, are not sensitive to $n$. Any variations within these ranges lead to reasonable performance. For E2E tasks, a looser filtering approach results in better text flexibility, leading to slight performance improvements.
Notably, arithmetic reasoning tasks exhibit unique behavior. Tighter filtering significantly improved the performance on the GSM8K, AddSub, and ASDiv datasets.

We believe that these differences in sensitivity arise from the nature of the tasks.
The outputs of tasks other than arithmetic reasoning exhibit a certain level of determinism (specific answers to questions, sentences conveying the same semantics in the target language, or restaurant reviews containing specific information). Hence, the output distributions of different models will demonstrate strong consistency. As illustrated in Table~\ref{table:ana2}, in the case of Chinese$\rightarrow$English translation task, models exhibit marginal differences in predicting the next token. As a result, the filtering strategy has minimal impact here. 
In contrast, arithmetic reasoning tasks generate a series of intermediate reasoning steps. Since the same answer can be derived from multiple distinct reasoning paths, the output tokens exhibit inconsistency. As shown in Table~\ref{table:ana2}, there is a significant semantic difference between the distributions of the next token in the GSM8K task. Employing tighter filtering here can effectively eliminate models generating unfaithful tokens.

\begin{figure}[!ht]
    \centering
    \includegraphics[width=\hsize]{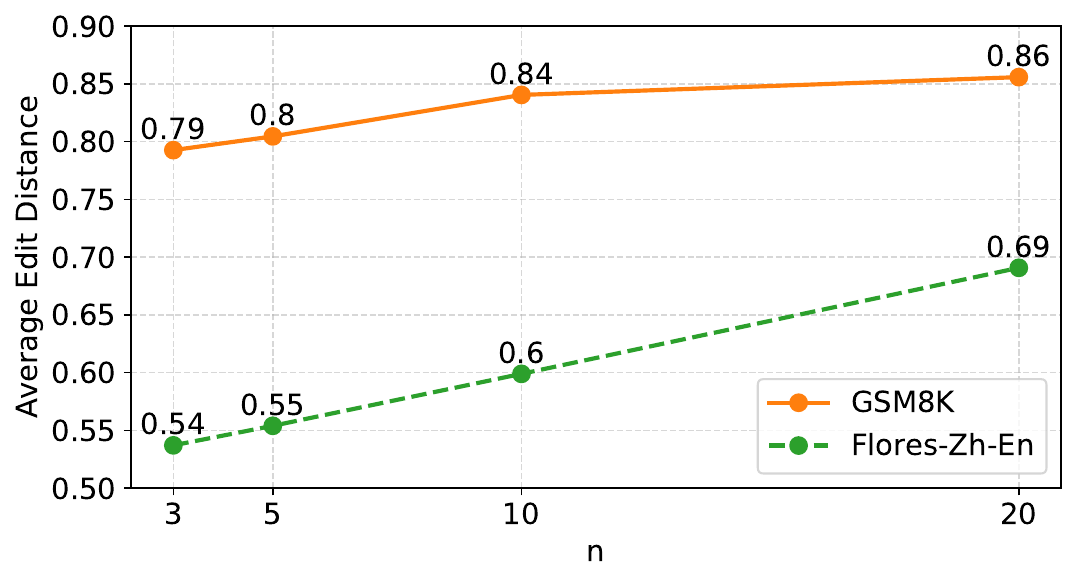}
    \caption{The average edit distance of GSM8K (orange solid line) and Flores-Zh-En (green dotted line) tasks across various top-$n$ ranges. The average edit distance indicates the output token diversity.}
    \label{fig.ana2}
\end{figure}

To verify our hypothesis, we conduct further experimental analysis on tasks with the highest sensitivity (GSM8K) and lowest sensitivity (Machine Translation).
Since tokens are very fine-grained units, spelling variations can directly represent semantic differences. Hence, We specifically define diversity as the average edit distance between the top-$n$ tokens and the top-1 token generated by a model.
We conducted a statistical analysis on the outputs at 10,000 positions in both datasets. As depicted in Figure~\ref{fig.ana2}, across various top-$n$ ranges, the edit distance for the GSM8K task consistently exceeds that of Flores, confirming our hypothesis.

\subsection{Effect of Number of Ensemble Models}
\begin{figure}[!t]
    \centering
    \includegraphics[width=\hsize]{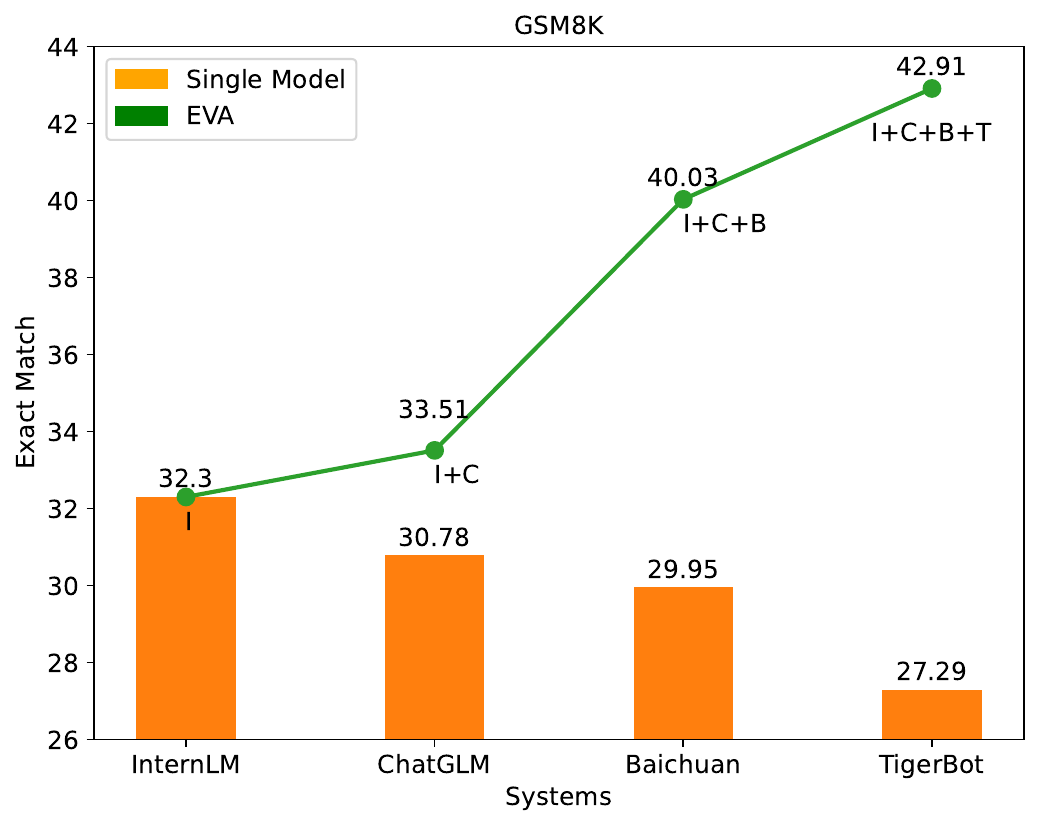}
    \caption{Effect of number of ensemble models. The orange bars represent the performance of individual models, while the green line denotes the result of ensembling multiple models, denoted by their initials.}
    \label{fig.ana3}
\end{figure}

As shown in Figure~\ref{fig.ana3}, we demonstrate the changes in ensemble performance on the GSM8K dataset as the number of ensemble models increases. We observe that even as the performance of newly added models gradually decreases, EVA consistently brings further improvements, which indicates that EVA effectively unleashes the complementary potential of different models by unifying the vocabulary space. Moreover, this confirms that different models possess distinct knowledge. The knowledge within underperforming models is not entirely covered by better-performing ones, leaving space for further enhancement via ensembling.

\section{Related Work}
\label{sec:bibtex}
Ensemble learning is a widely adopted technique to improve performance on a given task and provide robust generalization by leveraging multiple complementary systems~\citep{zhou2017neural,liu2018comparable,ganaie2022ensemble}. Existing ensemble methods can be divided into two categories: selection-based ensemble and generation-based ensemble.

\noindent\textbf{Selection-based Ensemble}
Selection-based ensemble methods select the best output from multiple outputs. \citet{shnitzer2023large} employs benchmark datasets to learn a router model responsible for selecting the best LLM out of a collection of models for a given task. FrugalGPT~\citep{chen2023frugalgpt} calls LLMs sequentially until a dedicated scoring model deems the generation acceptable to effectively and efficiently leverage different LLMs. \citet{ravaut2022summareranker};\citet{liu2021simcls};\citet{liu-etal-2022-brio} train dedicated scoring or ranking models for text summarization. \citet{farinhas2023empirical} demonstrated that minimum Bayes risk decoding is an effective ensemble method for LLM-based machine translation.

However, such methods are limited by the output quality of the candidate models and are unable to generate outputs superior to those of existing models. Nevertheless, the distinctions among candidates could be quite subtle. A model's output might outperform one part compared to another model's output yet lag behind in other parts. Selecting among existing answers limits the release of the complementary potential of the ensemble.

\noindent\textbf{Fusion-based Ensemble}
Compared to selection-based methods, fusion-based ensemble approaches bypass the limitation of existing complete outputs, often yielding superior outputs. \citet{llm-blender-2023} presents a general ensemble framework utilizing a pair ranker to filter the top K optimal outputs, followed by a fusion model to merge and generate the final output.  Furthermore,~\citet{izacard2021leveraging} enhances question answering by amalgamating retrieved text, while~\citet{ravaut2022towards} applies generative fusion methods to text summarization. However, a fusion model typically needs to have a size comparable to the base model. For instance,~\citet{llm-blender-2023} employs a 3B-sized model as a fusion model, significantly elevating the training and inference costs.

Our proposed EVA conducts fine-grained ensemble at each generation step, not only obtaining new results distinct from individual model outputs but also incurring almost negligible training costs for mapping vocabularies.
Furthermore, our approach exhibits strong performance without the need for training on specific task datasets, demonstrating excellent generalization capabilities.

\section{Conclusion}
In this paper, we propose a novel ensemble method named EVA, which effectively bridges the lexical gap between different LLMs and facilitates fine-grained ensemble at each generation step. Compared to ensemble methods that select or fuse completely generated results, EVA provides intermediate ensemble results to candidate models, enabling them to benefit from higher-quality output prefixes, thereby unleashing their complementary potentials. Experimental results on NLG tasks and reasoning tasks demonstrate the effectiveness of our approach, which significantly improves overall performance on various natural language processing tasks.

\section*{Limitation}
Due to the inherent nature of the ensemble, our approach, like previous ensemble methods, requires performing inference N times when ensembling N models. However, we want to argue that those inferences can be executed in parallel because they are totally independent.

\section*{Acknowledgments}
This work is supported by National Key R\&D Program of China 2022ZD0160602 and the Natural Science Foundation of China 62122088.

\bibliography{custom}

\newpage 
\appendix

\section{Effect of the Pivot Model}\label{sec:app3}
When the vocabulary is small, in order to avoid OOV problems, the granularity of word segmentation is relatively fine, and the tokens in the vocabulary are more likely to be high-frequency subwords (e.g., sub) that appear in many words. On the contrary, a larger vocabulary means that the tokens are more diverse and specialized. More tokens with specific meanings (e.g., subject) will appear in the vocabulary.

Suppose we want to ensemble two models A (large vocabulary) and B (small vocabulary). 
If A is the pivot model, at each decoding step, we are more likely to get a token with specific meaning (e.g., subject), and B can segment the prefixes into finer tokens in its own way (subject\ ->\ sub / je / ct).
If b is the pivot model, we are more likely to get a token with ambiguous meaning (e.g., sub), and the way A handles "sub" (sub\ ->\ s / ub) is different from the way it handles "subject" (subject\ ->\ subject). This may affect the performance of the model.

Based on the above considerations, we choose the model with the largest vocabulary as the pivot model in our method, rather than selecting the best-performing model. This is a practical and effective approach in real-world scenarios, and it doesn't require prior knowledge of individual model performance.

\section{Effectiveness of Vocabulary Projection}\label{sec:app1}

We observe the results of vocabulary projection between different models and analyze the relationship between similarity scores and projection phenomena. In Table~\ref{tab.alignment}, we illustrate the observed results using the projection from LLaMA2-7B-Chat~\citep{touvron2023llama} to Baichuan2-7B-Chat~\citep{baichuan2023baichuan2} as an example. 
For token pairs with similarity scores between 0.6 and 1, most of them are completely aligned.
It should be noted that some special tokens demonstrate high similarity but lack semantic meaning in their alignment, clustering around a similarity score of 0.77. 
As the similarity decreases to the range of 0.4 to 0.6, minor inconsistencies that do not affect semantics begin to appear, such as singular and plural forms, uppercase and lowercase distinctions. 
Furthermore, as the similarity reduces to 0.1 to 0.4, phenomena shift towards partial alignment and cross-lingual alignment.
When the similarity drops below 0.1, the majority of alignments are meaningless.
Overall, approximately 82\% of the vocabulary achieved meaningful mappings, indicating the effectiveness of our vocabulary projection.

\section{Datasets}\label{sec:app2}

\paragraph{GSM8K} is a multi-step arithmetic reasoning dataset~\citep{cobbe2021gsm8k}, consists of high quality linguistically diverse grade school math word problems created by human problem writers. Evaluation metrics are Accuracy.

\paragraph{AddSub} consists of addition–subtraction word problems\citep{hosseini2014learning}. Evaluation metrics are Accuracy.

\paragraph{ASDiv} is a diverse (in terms of both language patterns and problem types) English math word problem corpus\cite{miao2020diverse}. Evaluation metrics are Accuracy.

\paragraph{Natural Questions (NQ)} is a question answering dataset in which questions consist of real anonymized, aggregated queries issued to the Google search engine~\citep{kwiatkowski2019natural}. Following OpenCompass~\citep{2023opencompass}, we repurposed the validation set for testing purposes. Evaluation metrics are Exact Match.

\paragraph{TriviaQA} contains questions authored by trivia enthusiasts~\citep{joshi2017triviaqa}. 
Again, we use the validation as test. Evaluation metrics are Exact Match.

\paragraph{Flores101} is a widely used benchmark dataset for machine translation~\citep{flores101}. Here we use the Chinese-English split and English-Chinese split for evaluation. Evaluation metrics are BLEU~\citep{post-2018-call} and ChrF~\citep{popovic2015chrf}.

\paragraph{E2E} is a data-to-text dataset~\citep{novikova2017e2e}. The input is a set of key-value attribute pairs, and the output is a description of the restaurant. Evaluation metrics are ROUGE-L\footnote{\url{https://github.com/GrittyChen/NLG-evaluation}}.

\section{Integration of Models of Different Sizes}\label{sec:app4}
Since our method only operates on the model output distribution, it is not constrained by the model size and internal structure. Therefore, differences in model sizes only reflect variations in model performance.

In our main experiment, the performance differences between different models are already very representative. For example, there is a 12\% accuracy difference between the best model and the worst model on the TriviaQA task. For the Flores-En-Zh task, the difference is 7 BLEU scores. 

In addition, we conduct an ensemble experiment with difference sizes including 6b, 7b and 13b on the ASDiv task. As shown in the Table~\ref{table:model_size}, our method consistently delivers stable performance improvements.

\begin{table}[!ht]
\centering
\begin{tabular}{@{}lc@{}}
\toprule
\textbf{Model}       & \textbf{ASDiv} \\ \midrule
LLaMA2-7B-Chat       & 55.02          \\
ChatGLM2-6B          & 60.52          \\
Baichuan2-13B-Chat   & 67.80          \\
TigerBot-13B-Chat-V3 & 56.47          \\
EVA(\textit{ours})            & 71.52          \\ \bottomrule
\end{tabular}

\caption{\label{table:model_size} 
Ensemble results of models of different sizes on the ASDiv task.}
\end{table}

\section{Human Analysis of Token Alignment}\label{sec:app5}
We sample 300 tokens for each task, and conduct human analysis on the top-1 token before vocabulary mapping and the top-1 token after vocabulary mapping. The statistical results are shown in the Table~\ref{table:human_ana}. We observe that on average, about 95\% of the vocabulary mapping is correct. This demonstrates the effectiveness of our vocabulary mapping in achieving strong matching. We also analyze examples of incorrect matching, which mainly include the following types of errors:

\begin{itemize}
    \item Related but semantic drift, such as \textit{research} -> \textit{study}.
    \item Conjunction, such as \textit{of} -> \textit{in}.
    \item Match incorrectly, such as \textit{(} -> \textit{H}.
\end{itemize}

\begin{table*}[!ht]
\begin{tabular}{@{}lllll@{}}
\toprule
\textbf{Task} & \textbf{Match Correctly} & \textbf{Related but Semantic Drift} & \textbf{Conjunction} & \textbf{Match Incorrectly} \\ \midrule
NQ            & \ \ \ \ \ \ \ \ 99.33\%                  & \ \ \ \ \ \ \ \ \ \ \ \ \ \ \ \ 0.00\%                              & \ \ \ \ \ \ 0.33\%               & \ \ \ \ \ \ \ \ \ \ 0.33\%                     \\
TriviaQA      & \ \ \ \ \ \ \ \ 98.00\%                  & \ \ \ \ \ \ \ \ \ \ \ \ \ \ \ \ 0.00\%                              & \ \ \ \ \ \ 0.33\%               & \ \ \ \ \ \ \ \ \ \ 1.67\%                     \\
ASDiv         & \ \ \ \ \ \ \ \ 90.67\%                  & \ \ \ \ \ \ \ \ \ \ \ \ \ \ \ \ 0.33\%                              & \ \ \ \ \ \ 1.33\%               & \ \ \ \ \ \ \ \ \ \ 7.67\%                     \\
AddSub        & \ \ \ \ \ \ \ \ 93.00\%                  & \ \ \ \ \ \ \ \ \ \ \ \ \ \ \ \ 0.00\%                              & \ \ \ \ \ \ 0.67\%               & \ \ \ \ \ \ \ \ \ \ 6.33\%                     \\
GSM8K         & \ \ \ \ \ \ \ \ 93.00\%                  & \ \ \ \ \ \ \ \ \ \ \ \ \ \ \ \ 0.00\%                              & \ \ \ \ \ \ 0.67\%               & \ \ \ \ \ \ \ \ \ \ 6.33\%                     \\
E2E           & \ \ \ \ \ \ \ \ 95.33\%                  & \ \ \ \ \ \ \ \ \ \ \ \ \ \ \ \ 0.00\%                              & \ \ \ \ \ \ 1.33\%               & \ \ \ \ \ \ \ \ \ \ 3.33\%                     \\
Flores-Zh-En  & \ \ \ \ \ \ \ \ 94.67\%                  & \ \ \ \ \ \ \ \ \ \ \ \ \ \ \ \ 2.33\%                              & \ \ \ \ \ \ 0.67\%               & \ \ \ \ \ \ \ \ \ \ 2.33\%                     \\
Flores-En-Zh  & \ \ \ \ \ \ \ \ 95.00\%                  & \ \ \ \ \ \ \ \ \ \ \ \ \ \ \ \ 2.33\%                              & \ \ \ \ \ \ 0.33\%               & \ \ \ \ \ \ \ \ \ \ 2.33\%                     \\
Average       & \ \ \ \ \ \ \ \ 94.88\%                  & \ \ \ \ \ \ \ \ \ \ \ \ \ \ \ \ 0.62\%                              & \ \ \ \ \ \ 0.71\%               & \ \ \ \ \ \ \ \ \ \ 3.79\%                     \\ \bottomrule
\end{tabular}
\caption{\label{table:human_ana} 
Human analysis of token alignment.}
\end{table*}

\end{document}